\documentclass[a4paper,twoside]{article}

\usepackage{epsfig}
\usepackage{subcaption}
\usepackage{calc}
\usepackage{amssymb}
\usepackage{amstext}
\usepackage{amsmath}
\usepackage{amsthm}
\usepackage{multicol}
\usepackage{pslatex}
\usepackage{apalike}
\usepackage{algorithm2e}
\usepackage[bottom]{footmisc}

\usepackage{graphicx}
\usepackage{amsmath}
\usepackage{amssymb}
\usepackage{booktabs}
\usepackage{multirow} 
\usepackage{booktabs} 
\usepackage[table, svgnames, dvipsnames]{xcolor}
\usepackage{pgfplots}

\usepackage{SCITEPRESS}     

\pgfplotsset{compat=1.18}
\usepackage{xcolor}
\definecolor{darkbrown}{rgb}{0.36, 0.25, 0.20} 

\begin{document}

\title{Data-Driven Fairness Generalization for Deepfake Detection}

\author{\authorname{Uzoamaka Ezeakunne\sup{1}, Chrisantus Eze\sup{2} and Xiuwen Liu\sup{1}}
\affiliation{\sup{1} Department of Computer Science, Florida State University, Tallahassee FL, USA}
\affiliation{\sup{2}Department of Computer Science, Oklahoma State University, Stillwater OK, USA}
\email{ufe18@fsu.edu, chrisantus.eze@okstate.edu, liux@cs.fsu.edu}
}


\abstract{Despite the progress made in deepfake detection research, recent studies have shown that biases in the training data for these detectors can result in varying levels of performance across different demographic groups, such as race and gender. These disparities can lead to certain groups being unfairly targeted or excluded. Traditional methods often rely on fair loss functions to address these issues, but they under-perform when applied to unseen datasets, hence, fairness generalization remains a challenge. In this work, we propose a data-driven framework for tackling the fairness generalization problem in deepfake detection by leveraging synthetic datasets and model optimization. Our approach focuses on generating and utilizing synthetic data to enhance fairness across diverse demographic groups. By creating a diverse set of synthetic samples that represent various demographic groups, we ensure that our model is trained on a balanced and representative dataset. This approach allows us to generalize fairness more effectively across different domains. We employ a comprehensive strategy that leverages synthetic data, a loss sharpness-aware optimization pipeline, and a multi-task learning framework to create a more equitable training environment, which helps maintain fairness across both intra-dataset and cross-dataset evaluations. Extensive experiments on benchmark deepfake detection datasets demonstrate the efficacy of our approach, surpassing state-of-the-art approaches in preserving fairness during cross-dataset evaluation. Our results highlight the potential of synthetic datasets in achieving fairness generalization, providing a robust solution for the challenges faced in deepfake detection.}

\onecolumn \maketitle \normalsize \setcounter{footnote}{0} \vfill

\section{\uppercase{Introduction}}
\label{sec:intro}

\definecolor{darkgreen}{rgb}{0, 0.4, 0}

Deepfake technology, which combines "deep learning" and "fake," represents a significant advancement in media manipulation capabilities. Leveraging deep learning techniques, it enables highly convincing facial manipulations and replacements in digital media. While technologically impressive, this capability poses serious societal risks, particularly in spreading misinformation and eroding public trust. In response, researchers have developed various deepfake detection techniques that have shown promising accuracy rates in identifying manipulated content \cite{detect1,detect2,detect3,detect4,detect5,detect6}.

However, a critical challenge has emerged in the form of fairness disparities across demographic groups as described in \cite{unfair1,fairgen_27,data1,fairgen_30}. Studies have revealed that current detection methods perform inconsistently across different demographics, particularly showing higher accuracy rates for individuals with lighter skin tones compared to those with darker skin tones \cite{unfair1,dark1}. This bias creates a concerning vulnerability where malicious actors could potentially target specific demographic groups with deepfakes that are more likely to evade detection.

While recent algorithmic approaches have shown promise in improving detection fairness when training and testing data use similar forgery techniques \cite{fair}, the real-world application presents a more complex challenge. Deepfake detectors, typically developed and trained on standard research datasets, must ultimately operate in diverse real-world environments where they encounter deepfakes created using various, potentially unknown forgery techniques. This generalization capability is crucial for practical deployment, particularly on social media platforms where manipulated content proliferates.

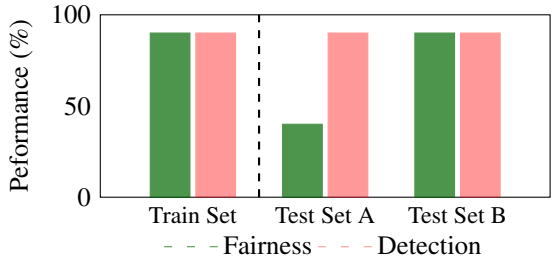
\begin{figure}[t]
\centering
\begin{tikzpicture}
\begin{axis}[
    ylabel={Peformance (\%)},
    ytick={0, 50, 100},
    ymin=0,
    ymax=100,
    bar width=0.3,
    width=7.5cm,
    height=4cm,
    xtick={1,2,3},
    xticklabels={{\footnotesize Train Set},{\footnotesize Test Set A},{\footnotesize Test Set B}},
    tick style={draw=none},
    legend style={
        draw = none,
        at={(0.5,-0.15)},
        anchor=north,
        legend columns=-1,
    },
    enlarge x limits=0.35
]
\addplot[
    style={darkgreen, fill=darkgreen, opacity=0.7},
    ybar=2*\pgflinewidth,
    bar shift=-0.3cm
] 
coordinates {(1,90) (2,40) (3,90)};

\addplot[
    style={red, fill=red, opacity=0.4},
    ybar=2*\pgflinewidth,
    bar shift=0.3cm
] 
coordinates {(1,90) (2,90) (3,90)};

\addplot[
    domain=0:10, 
    samples=2,
    style={dashed, thick, black}
]
coordinates {(1.5,0) (1.5,100)}; 

\legend{Fairness, Detection}


\end{axis}
\end{tikzpicture}
\caption{Fairness Generalization Comparison: A detector's accuracy can generalize well to unseen test sets A and B, maintaining consistent detection performance. However, while fairness metrics are preserved on test set B, they fail to generalize to test set A, highlighting the challenge of achieving consistent fairness across different datasets.}
\label{fig:gen}
\end{figure}

While current methods of deepfake detection focus almost entirely on detection accuracy, it is very critical to also consider how well these approaches generalize to fairness. As shown in Figure \ref{fig:gen}, if a detector was trained on a dataset and its performance for fairness and detection on the training set are shown on the left bars, and the same detector is tested on an unseen dataset (test set A) at the middle, the detection performance, mostly in terms of model accuracy may generalize but fairness may fail to generalize. On the other hand, when tested on another unseen dataset (test set B), it can retain its fairness and detection performance. As can be seen in Figure \ref{fig:gen}, the detector's detection performance can be generalized to both datasets but could fail to ensure fairness generalization on test set A.

Our research addresses these challenges through a comprehensive approach that combines data-centric and algorithmic solutions. We propose a novel framework that utilizes self-blended images (SBI) for synthetic data generation to balance demographic representation in the training data, addressing a key source of fairness disparities. This is coupled with a multi-task learning architecture that simultaneously optimizes for both detection accuracy and demographic fairness. The architecture employs an EfficientNet backbone for feature extraction, with separate heads for deepfake detection and demographic classification.

To enhance generalization capabilities, we implement Sharpness-Aware Minimization (SAM) optimization \cite{fairgen}, which helps find more robust solutions by flattening the loss landscape. Our framework also incorporates a carefully designed loss function that balances classification accuracy, demographic prediction, and fairness constraints. This approach represents a significant advancement in creating deepfake detection systems that are not only accurate but also fair and generalizable across different demographic groups and forgery techniques.

As demonstrated in our experimental results, while traditional detectors may maintain detection accuracy across different test sets, they often fail to preserve fairness when encountering new data. Our method addresses this limitation by specifically optimizing for both performance metrics, ensuring consistent fairness across different demographic groups even when tested on previously unseen datasets.

Therefore, our contributions are as follows:

\begin{itemize}

    \item We propose synthetic data balancing, to balance training datasets across demographics to aid fair learning in deepfake detection.
    
    \item Our framework introduces a multi-task architecture, leveraging two classification heads; one for deepfake detection and another for learning the demographic features of the datasets. This enables the system to not only detect forgery but also ensure that it is aware of demographic biases, making it more equitable across different groups.
    
    \item The multi-task learning approach, combined with sharpness-aware loss optimization and robust feature extraction, makes the system more effective at generalizing the fairness and detection performances to unseen datasets. 
    \item We performed intra-dataset and cross-dataset evaluations to show the improved performance of this method.
\end{itemize}

\section{Related Work}
\label{sec:related}

\subsection{Deepfake Detection}
To train deepfake detection models that perform well during training and in practice; on unseen datasets, several studies \cite{rw2_45,rw1_59,rw1_19,rw1_31} have introduced novel methods for manually synthesizing a variety of face forgeries similar to deepfakes. These methods help deepfake detection models learn more generalized representations of artifacts, improving their ability to identify deepfakes across different scenarios.

The authors in \cite{rw1_5} proposed enhancing the diversity of forgeries through adversarial training to improve the robustness of deepfake detectors in recognizing various forgeries. FaceCutout \cite{rw3_14} uses facial landmarks and randomly cuts out different parts of the face (such as the mouth, eye, etc.) to improve the robustness of deepfake detection. Face-Xray \cite{rw2_45} on the other hand, involves generating blended images (BI) by combining two different faces using a global transformation and then training a model to distinguish between real and blended faces. The authors in \cite{rw1_43} expanded on this concept by creating a synthetic training dataset with self-blended images (SBI), which are created by applying a series of data augmentation techniques to real images. SBI has demonstrated even better generalization to previously unseen deepfakes compared to Face-Xray \cite{rw2_45}. After synthesizing forged images (i.e., pseudo deepfakes) using specialized augmentation techniques, they trained a binary classification model to perform the detection task. 

\subsection{Fairness in Deepfake Detection}
Despite efforts to enhance the generalization of deepfake detection to unseen data, limited progress has been made in mitigating biased performance when testing on both the same (intra-domain) and different (cross-domain) dataset distributions.

Recent studies have highlighted significant fairness issues in deepfake detection, revealing biases in both datasets and detection models \cite{fairgen_30}. The work done by \cite{unfair1} and \cite{dark1} discovered substantial error rate differences across demographic groups, while \cite{fairgen_33} found that the MesoInception-4 model by \cite{meso} was biased against both genders. The authors in \cite{fairgen_27} advocated for diverse dataset annotations to address these biases, and \cite{data1} introduced a gender-balanced dataset to reduce gender-based performance bias. However, these techniques led to only modest improvements and required extensive data annotation. Furthermore, \cite{fair} worked to improve fairness within the same dataset (intra-domain), but did not address fairness generalization between different datasets (cross-domain). 

\subsection{Fairness Generalization in Deepfake Detection}
Very little work has been done to achieve fairness generation in deepfake detection. The recent work done in \cite{fairgen} proposed a novel feature-based technique to achieve fairness generalization and preservation. They combine disentanglement learning, fairness learning, and loss optimization to preserve fairness in deepfake detection. In the disentanglement learning module, they utilized a disentanglement loss to expose demographic and domain-agnostic forgery features. While the fairness learning module fused the two features from the disentanglement learning module to obtain predictions for the samples. Their framework, however, was trained on a dataset with imbalanced demographic groups which introduced some bias to their model.

Our framework approaches and solves fairness generalization using a data-driven approach. Our research prioritizes fairness and eliminates bias due to dataset imbalance by using synthetic datasets to balance the dataset distribution across the different groups. We also used a multi-task learning approach, combined with fairness and loss sharpness-aware optimization and robust feature extraction to achieve significant performance improvements on both intra-dataset evaluation and cross-dataset evaluation.

\section{Proposed Method}
\label{sec:proposed}
Our method aims to enhance both the generalization and fairness of deepfake detection models through a data-centric strategy. This section provides a detailed overview of the problem formulation, followed by the process of generating our synthetic dataset. We then outline our approach for addressing dataset imbalance and conclude with a discussion of our multi-task learning framework, designed to simultaneously optimize for accuracy and fairness across demographic groups.

\subsection{Problem Setup}
Given an input face image \( I \), the goal is to train a deepfake detection model \( g \) that classifies whether \( I \) is a deepfake (fake) or a real face. The model \( g \) is defined as:

\begin{equation}
g: I \to \{0, 1\}
\label{eq: prob1}
\end{equation}

where \( g(I) \) indicates the predicted label, with \( g(I) = 1 \) denoting a deepfake and \( g(I) = 0 \) denoting a real face.

\underline{\emph{Detection Accuracy: }} The primary goal is to maximize the detection accuracy. Given a set of face images \( \{I_i\}_{i=1}^N \) with true labels \( \{y_i\}_{i=1}^N \), the overall accuracy of the model \( g \) is:

\begin{equation}
\text{Accuracy} = \frac{1}{N} \sum_{i=1}^N \mathbf{1}(g(I_i) = y_i)
\label{eq: prob2}
\end{equation}

where \( \mathbf{1}(\cdot) \) is the indicator function that equals 1 if the prediction matches the true label.

\underline{\emph{Fairness Constraint: }} The detector should perform fairly across different demographic groups. Let \( D \) represent the demographic group associated with an image, and \( \mathcal{D}_k \) be the set of images from the demographic group \( k \). Define \( \text{Accuracy}_k \) as the accuracy of the detector on images from demographic group \( k \):

\begin{equation}
\text{Accuracy}_k = \frac{1}{|\mathcal{D}_k|} \sum_{i \in \mathcal{D}_k} \mathbf{1}(g(I_i) = y_i)
\label{eq: prob3}
\end{equation}

The fairness of the detector is assessed by minimizing the maximum disparity in accuracy across different demographic groups:

\begin{equation}
\text{Fairness} = \max_{k, l} \left| \text{Accuracy}_k - \text{Accuracy}_l \right|
\label{eq: prob4}
\end{equation}

where \( k \) and \( l \) denote different demographic groups. The objective is to minimize this fairness disparity.

\begin{figure*}[tb]
  \centering
   \includegraphics[width=1.0\linewidth]{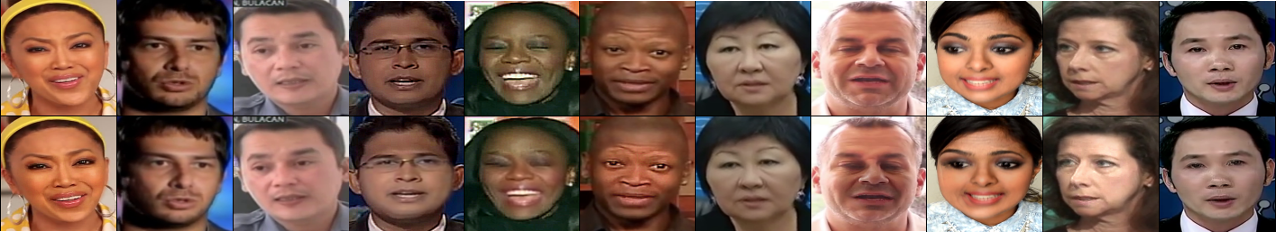}
   \caption{Sample of Facial Images. The real images (top row) and their corresponding fake/synthetic image
(bottom row).}
   \label{fig:sbi_images}
\end{figure*}

\subsection{Synthetic Data Balancing}
\label{sec:syn}
To enhance fairness in deepfake detection, we utilize a technique based on self-blended images (SBI), as proposed in \cite{rw1_43}. This technique involves generating synthetic images to balance the demographic distribution of the dataset, thereby improving fairness across various demographic groups. The dataset samples are shown in Figure \ref{fig:sbi_images}. Each real sample (upper row) has a corresponding fake sample (bottom row), so the dataset is also balanced across real and fake classes. The process of data balancing using a synthetic dataset is detailed in the sections below.

\subsubsection*{Generation of Synthetic Images}
\label{subsec:gensyn}
Let $\mathcal{I} = \{I_1, I_2, \ldots, I_n\}$ represent the set of real face images, where each $I_i$ belongs to a specific demographic group. The goal is to generate a set of self-blended images $\mathcal{S} = \{S_1, S_2, \ldots, S_m\}$ using the following steps:

\underline{\emph{Base Image Selection:}} Select a diverse subset of real face images, $\mathcal{I}_{base} \subset \mathcal{I}$, ensuring coverage across demographic categories.
    
\underline{\emph{Augmentation Process:}} Apply a series of transformations to generate self-blended images. For a base image $I_i$, the synthetic image $S_j$ is generated using a blend function $B$ as follows:
    \begin{equation}
    S_j = B(I_i, T_k(I_i))
    \label{eq:syn1}
     \end{equation}
where $T_k$ represents a transformation operation such as scaling, rotation, or color adjustment, and $B$ is the blending function that combines $I_i$ with $T_k(I_i)$.
\subsubsection*{Dataset Balancing} To balance the dataset, we increase the representation of underrepresented demographic groups:

\underline{\emph{Demographic Representation:}} Define the demographic groups $\mathcal{D} = \{D_1, D_2, \ldots, D_r\}$. Let $\mathcal{I}_{D_k} \subset \mathcal{I}$ be the set of images belonging to demographic group $D_k$. To balance the dataset, we create synthetic images $\mathcal{S}_{D_k}$ for each group $D_k$:

\begin{equation}
\mathcal{B} = \mathcal{I} \cup \mathcal{S}
\label{eq:syn2}
\end{equation}

where $\mathcal{B}$ is the balanced dataset and $\mathcal{S} = \bigcup_{k=1}^{r} \mathcal{S}_{D_k}$. 

This dataset $\mathcal{B}$, is balanced across all the demographics, and also equal in terms of the number of real and fake samples.

\subsection{Multi-task Learning}
\begin{figure*}[tb] 
\centering
\includegraphics[scale=0.4]{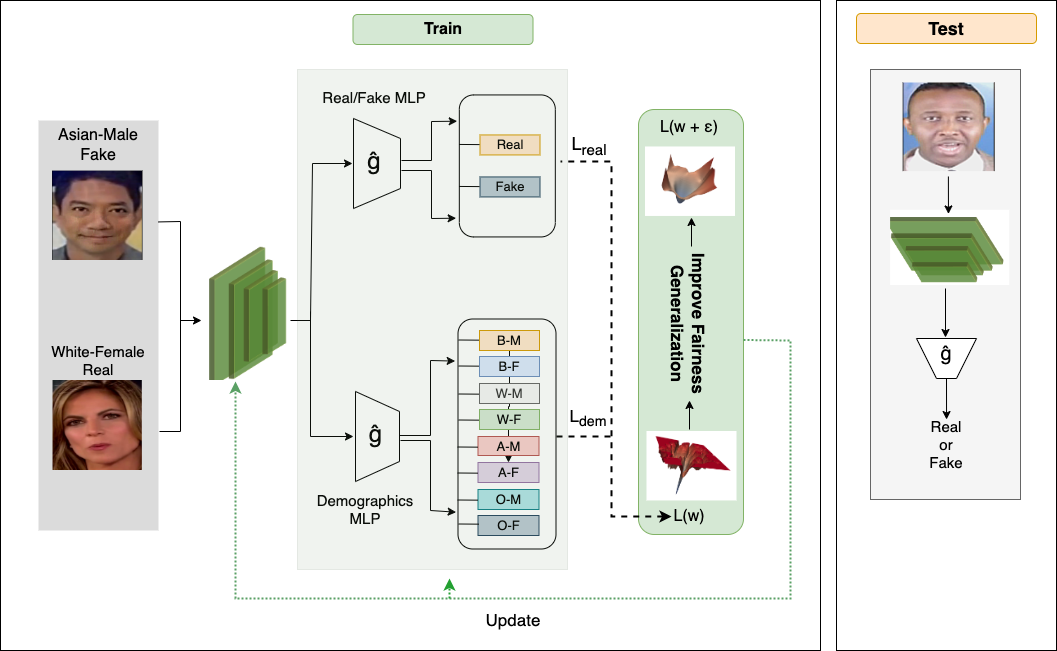} 
   \caption{An overview of our proposed method. We utilize EfficientNet \cite{eff} to extract deep features from input images for the feature extraction module. For the classification module, two heads are used: the real/fake head predicts whether the input is real or fake, and the demographic head predicts the demographic group. The demographic classification head outputs probabilities for one of eight demographic categories based on both gender and ethnicity. These categories include Black-Male (B-M), Black-Female (B-F), White-Male (W-M), White-Female (W-F), Asian-Male (A-M), Asian-Female (A-F), Other-Male (O-M), and Other-Female (O-F). We use SAM (Sharpness-Aware Minimization) for the optimization module to flatten the loss landscape and enhance fairness generalization across demographic groups. }

   \label{fig:arch}
\end{figure*}

We propose a dual-task learning framework to enhance the fairness of deepfake detection models by leveraging demographic information. The architecture shown in Figure \ref{fig:arch}, is designed to improve the generalization of deepfake detection across various demographic groups, reducing bias in prediction accuracy between these groups.

The input images are passed through the EfficientNet encoder \cite{eff}, which extracts high-dimensional feature representations. These features capture both low-level information (like textures and edges) as well as high-level semantic features (such as facial structure and potential tampering artifacts). The extracted features serve as the input for the real/fake classification head and the demographic head.

\underline{\emph{Real/Fake Classification Head:}}  This head consists of a Multi-Layer Perceptron (MLP) that processes the extracted features and predicts whether the input image is real or fake.
The output of this MLP is a probability score indicating the likelihood that the image is a deepfake or a genuine one. 
This branch focuses on learning the subtle cues and artifacts associated with the deepfake generation, such as inconsistencies in lighting, facial details, or blurring artifacts around the face.

\underline{\emph{Demographic Head:}}
In parallel to the real/fake classification, the features are passed through a second MLP for demographic group classification. This MLP is designed to categorize the input into one of eight demographic groups, such as Black-Male (B-M), White-Female (W-F), Asian-Male (A-M), etc.
The purpose of this head is to ensure that the model learns demographic-specific features that are independent of deepfake artifacts. These features are crucial for enhancing the fairness of the model across diverse demographic groups.

\subsubsection*{Loss Function}
We define the loss function to incorporate both the standard classification loss and a fairness penalty. Specifically, the total loss function \( \mathcal{L} \) is composed of three terms: the classification loss, the demographic loss, and the fairness penalty.

\underline{\emph{Classification Loss: }}
   Let \( \textbf{p} = [p_1, p_2, \dots, p_N] \) be the predicted probabilities and \( \textbf{y} = [y_1, y_2, \dots, y_N] \) be the true binary labels. The binary cross-entropy loss \( \mathcal{L}_{\text{real}} \) is defined as:

 \begin{equation}
   \mathcal{L}_{\text{real}} = - \frac{1}{N} \sum_{i=1}^{N} \left[ y_i \log(p_i) + (1 - y_i) \log(1 - p_i) \right]
    \label{eq:loss1}
 \end{equation}

   where \( N \) is the number of samples, \( y_i \) is the true binary label (0 or 1) for the \( i \)-th sample, and \( p_i \) is the predicted probability for the positive class.

\underline{\emph{Demographic Loss: }}
   Let \( \textbf{p}_{\text{dem},i} = [p_{\text{dem},i,1}, \dots, p_{\text{dem},i,8}] \) represent the predicted probabilities for the \( i \)-th sample across 8 demographic groups, and \( \textbf{y}_{\text{dem},i} = [y_{\text{dem},i,1}, \dots, y_{\text{dem},i,8}] \) be the one-hot encoded true labels for those groups. The demographic loss \( \mathcal{L}_{\text{dem}} \) is:

\begin{equation}
   \mathcal{L}_{\text{dem}} = - \frac{1}{N_{\text{dem}}} \sum_{i=1}^{N_{\text{dem}}} \sum_{c=1}^{8} y_{\text{dem}, i, c} \log(p_{\text{dem}, i, c})
   \label{eq:loss2}
\end{equation}

   where \( N_{\text{dem}} \) is the number of samples in the demographic component, and the summation runs over the 8 demographic groups.

\underline{\emph{Fairness Penalty: }}
   To quantify fairness, we compute the variance of the accuracy rates across different demographic groups. Let \( \textbf{a}_k \) denote the accuracy for demographic group \( k \). The variance of accuracies \( \text{Var}_{\text{acc}} \) is:

\begin{equation}
   \text{Var}_{\text{acc}} = \frac{1}{K} \sum_{k=1}^K \left( \text{Accuracy}_k - \bar{\text{Accuracy}} \right)^2
   \label{eq:loss3}
\end{equation}

   where \( K \) is the number of demographic groups, \( \text{Accuracy}_k \) is the accuracy for group \( k \), and \( \bar{\text{Accuracy}} \) is the mean accuracy across all groups.

\underline{\emph{Total Loss: }}
   The total loss function \( \mathcal{L} \) is a combination of the standard loss, the fairness penalty, and the fairness loss. The total loss is defined as:

\begin{equation}
   \mathcal{L} = \mathcal{L}_{\text{real}} + \lambda \cdot \text{Var}_{\text{acc}} + \mathcal{L}_{\text{dem}}
   \label{eq:loss4}
\end{equation}

where $\lambda$  is a hyperparameter controlling the weight of the fairness penalty.

\subsubsection*{Optimization}
We employ the Sharpness-Aware Minimization (SAM) optimizer \cite{sam} as our optimizer. SAM is designed to improve the generalization performance of deep neural networks by encouraging the model to find parameter spaces that are not only good at minimizing the training loss but also robust to small perturbations in the weight space. This results in a flatter and smoother loss landscape, which has been shown to correlate with better generalization.

The SAM optimizer modifies the traditional gradient-based update by introducing a two-step process that seeks to find parameters that minimize both the loss and its sensitivity to perturbations. Let \( \mathcal{L}(\mathbf{w}) \) denote the total loss function of the model, where \( \mathbf{w} \) represents the network's parameters. SAM performs the following steps:

 \underline{\emph{Perturbation Step: }}
    SAM first finds a perturbation \( \epsilon \) that maximizes the loss function around the current parameters. The perturbation \( \epsilon \) is chosen to make the loss worse within a neighborhood of the current parameter values:
    \begin{equation}
    \epsilon(\mathbf{w}) = \arg \max_{\|\epsilon\| \leq \rho} \mathcal{L}(\mathbf{w} + \epsilon)
    \label{eq:opt1}
    \end{equation}
    where \( \rho \) controls the size of the neighborhood in which the optimizer searches for the sharpest directions.
    
\underline{\emph{Parameter Update Step: }}
    Once the worst-case perturbation \( \epsilon \) is found, the SAM optimizer updates the parameters by minimizing the loss function at \( \mathbf{w} + \epsilon \):
    \begin{equation}
    \mathbf{w} \leftarrow \mathbf{w} - \eta \nabla \mathcal{L}(\mathbf{w} + \epsilon)
    \label{eq:opt2}
    \end{equation}
    where \( \eta \) is the learning rate.

This two-step process leads to solutions in flatter regions of the loss surface, improving the model's generalization and robustness.


\section{Experiments}
\label{sec:exper}

\begin{figure*}[tb]
    \centering
    \begin{minipage}{0.45\textwidth}
        \centering
        \begin{tikzpicture}
        \begin{axis}[
            grid=major,
            grid style={line width=.2pt, draw=gray!50},
            ylabel={Accuracy (\%)},
            ylabel style={at={(axis description cs:-0.1,.5)},anchor=south}, 
            ytick={70,80,90,100},
            ymin=70, ymax=105,
            bar width=0.25,
            width=7.5cm, height=4cm,
            xtick={1,2,3,4},
            xticklabels={{\scriptsize Xception},{\scriptsize DAW-FDD},{\scriptsize Lin et al.},{\scriptsize Ours}},
            tick style={draw=none},
            legend style={draw=none, at={(0.5,-0.15)}, anchor=north, legend columns=-1},
            enlarge x limits=0.3
        ]
        \addplot[
            style={blue, fill=blue, opacity=0.7},
            ybar,
            bar shift=-0.2cm
        ] 
        coordinates {(1,87.03) (2,92.65) (3,92.07) (4,92.39)};
        
        \addplot[
            style={red, fill=red, opacity=0.4},
            ybar,
            bar shift=0.2cm
        ] 
        coordinates {(1,88.61) (2,95.55) (3,95.94) (4,92.27)};
        
        \legend{\footnotesize{Male}, \footnotesize{Female}}
        \end{axis}
        \end{tikzpicture}
        \caption{Intra-dataset evaluation across different methods. Methods are trained and tested on the same distribution as the training set (FF++ \cite{ff++}): Performance comparison across gender. See Table \ref{tab:intracomparison} for exact numerical values}
        \label{fig:intragender}
    \end{minipage}\hfill
    \begin{minipage}{0.45\textwidth}

        \centering
        \begin{tikzpicture}
        \begin{axis}[
            grid=major,
            grid style={line width=.2pt, draw=gray!50},
            ylabel={Accuracy (\%)},
            ylabel style={at={(axis description cs:-0.1,.5)},anchor=south}, 
            ytick={70,80,90,100},
            ymin=70, ymax=105,
            bar width=0.25,
            width=7cm, height=4cm,
            xtick={0,1.5,3,4.5},
            xticklabels={{\scriptsize Xception},{\scriptsize DAW-FDD},{\scriptsize Lin et al.},{\scriptsize Ours}},
            tick style={draw=none},
            legend style={draw=none, at={(0.5,-0.15)}, anchor=north, legend columns=-1},
            enlarge x limits=0.3
        ]
        \addplot[
            style={darkbrown, fill=darkbrown, opacity=1.0},
            ybar,
            bar shift=-0.3cm
        ] 
        coordinates {(0,89.00) (1.5,94.67) (3,94.66) (4.5,92.04)};
        
        \addplot[
            style={yellow, fill=yellow, opacity=0.6},
            ybar,
            bar shift=-0.1cm
        ] 
        coordinates {(0,88.84) (1.5,94.29) (3,94.09) (4.5,92.28)};
        
        \addplot[
            style={orange, fill=orange, opacity=0.7},
            ybar,
            bar shift=0.1cm
        ] 
        coordinates {(0,83.72) (1.5,93.04) (3,93.46) (4.5,92.75)};
        
        \addplot[
            style={purple, fill=purple, opacity=0.9},
            ybar,
            bar shift=0.3cm
        ] 
        coordinates {(0,84.94) (1.5,94.68) (3,94.98) (4.5,92.25)};
        
        \legend{\footnotesize{Black}, \footnotesize{White}, \footnotesize{Asian}, \footnotesize{Others}}
        \end{axis}
        \end{tikzpicture}
        \caption{Intra-dataset evaluation across different methods. Methods are trained and tested on the same distribution as the training set (FF++ \cite{ff++}): Performance comparison across race. See Table \ref{tab:intracomparison} for exact numerical values}
        \label{fig:intrarace}
    \end{minipage}
\end{figure*}

\subsection{Experimental Settings}

\subsubsection*{Dataset}
To evaluate the fairness generalization capability of our proposed approach, we train our model on the widely used FaceForensics++ (FF++) dataset \cite{ff++}. For testing, we use FF++, Deepfake Detection Challenge (DFDC) \cite{dfdc}, and Celeb-DF \cite{celebdf}. We used the real images from these datasets and created synthetic images as the fake images using the method in Section \ref{sec:syn}. 

Since the original datasets do not include demographic information, we follow the approach in \cite{fair} for data processing and annotation. The datasets are categorized by race and gender, forming the following intersection groups: Black-Male (B-M), Black-Female (B-F), White-Male (W-M), White-Female (W-F), Asian-Male (A-M), Asian-Female (A-F), Other-Male (O-M), and Other-Female (O-F). The Celeb-DF \cite{celebdf} does not contain the Asian sub-group.

\subsubsection*{Evaluation Metrics}
For evaluating our approach, we use several performance metrics, including the Area Under the Curve (AUC), Accuracy, and True Positive Rate (TPR). These metrics allow us to benchmark our method against previous works. To assess the degree of fairness of the model, we analyze the accuracy gaps across different races and genders.

\subsubsection*{Baseline Methods}
For intra-dataset evaluation, we compare our approach against recent works in deepfake detection fairness, specifically DAW-FDD (2023) \cite{fair} and Lin et al. (2024) \cite{fairgen}. For cross-dataset evaluation, our method is compared to the fairness generalization method proposed by Lin et al. (2024) \cite{fairgen}. To the best of our knowledge, Lin et al. (2024) \cite{fairgen} is the prior work addressing the fairness generalization problem. We also used the Xception \cite{xcept} as a baseline for comparison during both intra-dataset evaluation and cross-dataset evaluation. The Xception network \cite{xcept} was implemented using transfer learning to perform the deepfake detection task.

\subsubsection*{Implementation Details}
All experiments are conducted using PyTorch. For our method, we set the batch size to 16 and train the model for 100 epochs. To incorporate fairness, we use a fairness penalty weight $\lambda$ of 20 (See Eq. \ref{eq:loss4}). For optimization, we employ the SAM optimizer \cite{sam}. The learning rate is set to 5e-4, momentum is configured at 0.9, and weight decay is set to \(5 \times 10^{-3}\). 

\subsection{Experimental Results}
In this section, we discuss the results of our evaluations and present our findings. We start by discussing the results of our intra-dataset evaluations and then, the results of the cross-dataset evaluations.

\subsubsection*{Intra-dataset Evaluation Results}

\begin{table*}[tb]

\centering
\setlength{\tabcolsep}{2pt}
\renewcommand{\arraystretch}{1.1}
{\small{
\begin{tabular}{@{\hskip 0pt}c|c|c|cccc@{\hskip 2pt}}
\hline

\multicolumn{3}{c|}{} &  \multicolumn{4}{c}{FF++ \cite{ff++}} \\
\cline{4-7}
 \multicolumn{3}{c|}{} & Xception  & DAW-FDD  & Lin et al.  & \cellcolor{gray!30}Ours \\
 \multicolumn{3}{c|}{} & \cite{xcept} & \cite{fair} & \cite{fairgen} & \cellcolor{gray!30}\\
 \hline
 
\multirow{6}{*}{Gender} & \multirow{3}{*}{Male}& TPR (\%)&\textbf{98.62} &93.48 & 92.13&\cellcolor{gray!30}91.91 \\
  & & AUC (\%) &90.23&96.91 & 97.56&\cellcolor{gray!30}\textbf{97.71}  \\
  & & ACC (\%) &87.03&\textbf{92.65} & 92.07&\cellcolor{gray!30}92.39 \\
  \cline{2-7}
  &\multirow{3}{*}{Female}& TPR (\%) &\textbf{98.85} &97.15 & 98.02&\cellcolor{gray!30} 91.89\\
  & & AUC (\%)  &93.48&98.05 & \textbf{98.94}&\cellcolor{gray!30}97.72 \\
  & & ACC (\%)  &88.61& 95.55& \textbf{95.94}&\cellcolor{gray!30} 92.27\\

\hline

\multirow{12}{*}{Race} & \multirow{3}{*}{Black}& TPR (\%) &\textbf{99.59} &96.36 & 97.17&\cellcolor{gray!30}91.27 \\
  & & AUC (\%) &95.85 &97.92 & \textbf{98.22} & \cellcolor{gray!30}97.57 \\
  & & ACC (\%)  & 89.00&\textbf{94.67} & 94.66 & \cellcolor{gray!30}92.04 \\
  \cline{2-7}
  &\multirow{3}{*}{White}& TPR (\%)  &\textbf{98.71}& 95.47 & 94.81 & \cellcolor{gray!30}91.88 \\
  & & AUC (\%) &92.57 & 97.54 & \textbf{98.21} & \cellcolor{gray!30}97.70 \\
  & & ACC (\%) &88.84 & \textbf{94.29} & 94.09 & \cellcolor{gray!30}92.28 \\
 \cline{2-7}
  &\multirow{3}{*}{Asian}& TPR (\%)  &\textbf{98.73}& 94.40 & 96.44 &\cellcolor{gray!30} 92.71\\
  & & AUC (\%) &89.40 & 96.39 & 97.54 & \cellcolor{gray!30} \textbf{97.91} \\
  & & ACC (\%) &83.72 & 93.04 & \textbf{93.46} & \cellcolor{gray!30}92.75 \\
\cline{2-7}
  &\multirow{3}{*}{Other}& TPR (\%) &\textbf{98.51 }& 95.77& 95.83 & \cellcolor{gray!30} 91.73\\
  & & AUC (\%)  & 89.87& 98.23& \textbf{98.58} & \cellcolor{gray!30}97.66 \\
  & & ACC (\%)  & 84.94&94.68 & \textbf{94.98} & \cellcolor{gray!30} 92.25\\

\hline

\end{tabular}
\caption{Intra-dataset evaluation compares different deepfake detection approaches by training and testing on the same dataset's distribution. Results are computed on each dataset's test set, with the best performance shown in bold.
}
\label{tab:intracomparison}
}}
\end{table*}


We perform intra-dataset evaluations to measure the model's performance on the same dataset distribution, assessing its ability to fit the training data. As shown in Table \ref{tab:intracomparison}, our method achieves comparable performance to baseline approaches in deepfake detection. However, it demonstrates a significant improvement over baseline methods in fairness preservation across demographic groups. Figure \ref{fig:intragender} presents the evaluation results for the gender demographic group, while Figure \ref{fig:intrarace} illustrates the results for the racial demographic group. As evidenced in these figures, the disparity between the Male and Female subgroups is substantially reduced in our approach with a 0.12\% difference (in accuracy) between the subgroups, compared to the 1.58\%, 2.90\% and 3.87\% differences between the subgroups for the baseline approaches. Additionally, the disparities among racial groups are significantly lower with our approach with a 0.71\% difference (which happens to be between asian and black - considering minimum and maximum accuracies) between the subgroups, while the baselines had differences of 5.28\%, 1.64\% and 1.52\%. These results highlight the effectiveness and superiority of our approach in achieving fairness generalization.

\subsubsection{Cross-dataset Evaluation Results}

\begin{table*}[tb]
\centering
\setlength{\tabcolsep}{2pt}
\renewcommand{\arraystretch}{1.1}
{\small{
\begin{tabular}{@{\hskip 0pt}c|c|c|ccc|ccc@{\hskip 2pt}}
\hline

\multicolumn{3}{c|}{} &  \multicolumn{3}{c}{DFDC \cite{dfdc}} & \multicolumn{3}{c}{Celeb-DF \cite{celebdf}} \\

\cline{4-9}
 \multicolumn{3}{c|}{}  & Xception & Lin et al.  & \cellcolor{gray!30}Ours &  Xception  & Lin et al.  & \cellcolor{gray!30}Ours \\
\multicolumn{3}{c|}{}  & \cite{xcept} & \cite{fairgen} & \cellcolor{gray!30} & \cite{xcept} & \cite{fairgen} & \cellcolor{gray!30}
 \\
 \hline
 
\multirow{6}{*}{Gender} & \multirow{3}{*}{Male}& 
     TPR (\%) & \textbf{85.11}& 37.81&\cellcolor{gray!30}35.49 & \textbf{98.68} & 76.89&\cellcolor{gray!30} 78.38\\
  & & AUC (\%) &59.97 &60.08 &\cellcolor{gray!30}\textbf{63.30} &36.97 & 66.93&\cellcolor{gray!30} \textbf{75.44} \\
  & & ACC (\%) & 43.75 &63.11 &\cellcolor{gray!30}\textbf{66.98} & 43.29& 56.16&\cellcolor{gray!30} \textbf{66.41}\\
  \cline{2-9}
  &\multirow{3}{*}{Female}& 
     TPR (\%) &\textbf{77.59} &44.36 &\cellcolor{gray!30}40.57 &\textbf{ 99.24}& 90.19&\cellcolor{gray!30} 86.67\\
  & & AUC (\%) &50.70 & \textbf{60.06}&\cellcolor{gray!30} 59.12& 53.26& 75.83&\cellcolor{gray!30}\textbf{82.09}\\
  & & ACC (\%) &45.60 &58.61 &\cellcolor{gray!30} \textbf{58.98}& \textbf{83.95}& 81.63&\cellcolor{gray!30}81.88\\
  \hline

\multirow{12}{*}{Race} & \multirow{3}{*}{Black}& 
     TPR (\%) & \textbf{76.28} &45.14 &\cellcolor{gray!30} 57.58&\textbf{98.68} & 76.89&\cellcolor{gray!30} 78.38\\
  & & AUC (\%) & 50.40 &59.81 &\cellcolor{gray!30}\textbf{68.83} &47.63 & 67.03&\cellcolor{gray!30}\textbf{72.53}\\
  & & ACC (\%) & 39.23& 58.66&\cellcolor{gray!30} \textbf{65.54}& 63.99& 64.10&\cellcolor{gray!30} \textbf{69.61}\\
  \cline{2-9}
  &\multirow{3}{*}{White}
    & TPR (\%) & \textbf{80.82} & 38.18&\cellcolor{gray!30}32.98 &\textbf{99.24} & 90.18&\cellcolor{gray!30} 86.69 \\
  & & AUC (\%) & 56.45&60.14 &\cellcolor{gray!30} \textbf{60.18}&53.58 & 77.02&\cellcolor{gray!30} \textbf{82.20}\\
  & & ACC (\%) & 48.79 &59.80 &\cellcolor{gray!30}\textbf{60.24} & \textbf{81.38}& 80.43&\cellcolor{gray!30} 81.15\\
\cline{2-9}
  &\multirow{3}{*}{Asian}&
     TPR (\%) & \textbf{82.50}&42.50 &\cellcolor{gray!30} 19.16&- & - &\cellcolor{gray!30}- \\
  & & AUC (\%) &54.75 &\textbf{51.78} &\cellcolor{gray!30} 43.89&- & - &\cellcolor{gray!30}- \\
  & & ACC (\%) & 22.38& 57.04&\cellcolor{gray!30} \textbf{71.39}& -& - &\cellcolor{gray!30}- \\
  \cline{2-9}
  &\multirow{3}{*}{Others}& 
     TPR (\%) & \textbf{87.68}&55.93 &\cellcolor{gray!30} 36.18 &\textbf{100} & 96.66&\cellcolor{gray!30}73.33 \\
  & & AUC (\%) &60.39 & \textbf{72.52}&\cellcolor{gray!30} 65.38&28.31 & 64.53&\cellcolor{gray!30}\textbf{84.97} \\
  & & ACC (\%) & 46.91& \textbf{71.43} &\cellcolor{gray!30} 65.82& 17.22& 32.78&\cellcolor{gray!30} \textbf{70.00}\\
\hline

\end{tabular}
\caption{Cross-dataset evaluation across different methods for deepfake detection. The methods were trained on (FF++ \cite{ff++}) and tested on DFDC \cite{dfdc} and Celeb-DF \cite{celebdf}.}
\label{tab:crosscomparison}
}}
\end{table*}

\begin{figure*}[ht]
    \centering
    \begin{minipage}{0.45\textwidth}
        \centering
\begin{tikzpicture}
\begin{axis}[
    grid=major, 
    grid style={line width=.2pt, 
    draw=gray!50}, 
    ylabel={Accuracy (\%)},
    xtick=\empty,
    ytick={0,10,20,30,40,50,60,70,80,90,100},
    ymin=30,
    ymax=80,
    bar width=0.3,
    width=7.5cm,
    height=4cm,
    xtick={1,2,3},
    xticklabels={{\footnotesize Xception},{\footnotesize Lin et al.},{\footnotesize Ours}},
    tick style={draw=none},
    legend style={
        draw = none,
        at={(0.5,-0.15)},
        anchor=north,
        legend columns=-1,
    },
    enlarge x limits=0.35
]
\addplot[
    style={blue, fill=blue, opacity=0.7},
    ybar=2*\pgflinewidth,
    bar shift=-0.3cm
] 
coordinates { (1,43.75) (2,63.11) (3,66.98)};

\addplot[
    style={red, fill=red, opacity=0.4},
    ybar=2*\pgflinewidth,
    bar shift=0.3cm
] 
coordinates {(1,45.60) (2,58.61) (3,58.98) };

\legend{Male, Female}

\end{axis}
\end{tikzpicture}
\caption{Cross-dataset evaluation across different methods. Methods are trained on (FF++ \cite{ff++}) and tested on DFDC \cite{dfdc}. Performance comparison is across gender. See Table \ref{tab:crosscomparison} for exact numerical values}
\label{fig:crossgenderdfdc}
    \end{minipage}\hfill
    \begin{minipage}{0.45\textwidth}
        \centering
\begin{tikzpicture}
\begin{axis}[
    grid=major, 
    grid style={line width=.2pt, 
    draw=gray!50}, 
    ylabel={Accuracy (\%)},
    xtick=\empty,
    ytick={0,10,20,30,40,50,60,70,80,90,100},
    ymin=20,
    ymax=80,
    bar width=0.2,
    width=7cm,
    height=4cm,
    xtick={0.3,1.5,2.7},
    xticklabels={ {\footnotesize Xception},{\footnotesize Lin et al.},{\footnotesize Ours}},
    tick style={draw=none},
    legend style={
        draw = none,
        at={(0.5,-0.15)},
        anchor=north,
        legend columns=-1,
    },
    enlarge x limits=0.35
]
\addplot[
    style={darkbrown, fill=darkbrown, opacity=1.0},
    ybar=2*\pgflinewidth,
    bar shift=-0.01cm
] 
coordinates {(0.0,39.23)(1.2,58.66) (2.4,65.54) };

\addplot[
    style={yellow, fill=yellow, opacity=0.6},
    ybar=2*\pgflinewidth,
    bar shift=0.01cm
] 
coordinates {(0.2,48.79) (1.4,59.80) (2.6,60.24) };

\addplot[
    style={orange, fill=orange, opacity=0.7},
    ybar=2*\pgflinewidth,
    bar shift=0.01cm
] 
coordinates {(0.4, 22.38)(1.6,57.04) (2.8,71.39) };

\addplot[
    style={purple, fill=purple, opacity=0.9},
    ybar=2*\pgflinewidth,
    bar shift=0.01cm
] 
coordinates {(0.6,46.91) (1.8,71.43) (3.0,65.82) };

\legend{\footnotesize{Black}, \footnotesize{White}, \footnotesize{Asian}, \footnotesize{Others}}


\end{axis}
\end{tikzpicture}
\caption{Cross-dataset evaluation across different methods. Methods are trained on (FF++ \cite{ff++}) and tested on DFDC \cite{dfdc}. Performance comparison is across race. See Table \ref{tab:crosscomparison} for exact numerical values}
\label{fig:crossracedfdc}
    \end{minipage}
\end{figure*}

\begin{figure*}[tb]
    \centering
    \begin{minipage}{0.45\textwidth}
        \centering
\begin{tikzpicture}
\begin{axis}[
    grid=major, 
    grid style={line width=.2pt, 
    draw=gray!50}, 
    ylabel={Accuracy (\%)},
    xtick=\empty,
    ytick={0,10,20,30,40,50,60,70,80,90,100},
    ymin=30,
    ymax=90,
    bar width=0.3,
    width=7.5cm,
    height=4cm,
    xtick={1,2,3},
    xticklabels={{\footnotesize Xception},{\footnotesize Lin et al. },{\footnotesize Ours}},
    tick style={draw=none},
    legend style={
        draw = none,
        at={(0.5,-0.15)},
        anchor=north,
        legend columns=-1,
    },
    enlarge x limits=0.35
]
\addplot[
    style={blue, fill=blue, opacity=0.7},
    ybar=2*\pgflinewidth,
    bar shift=-0.3cm
] 
coordinates {(1,43.29)(2,56.16) (3,66.41)};

\addplot[
    style={red, fill=red, opacity=0.4},
    ybar=2*\pgflinewidth,
    bar shift=0.3cm
] 
coordinates {(1,83.95) (2,81.63) (3,81.88)};

\legend{Male, Female}

\end{axis}
\end{tikzpicture}
\caption{Cross-dataset evaluation across different methods. Methods are trained on (FF++ \cite{ff++}) and tested on Celeb-DF \cite{celebdf}. Performance comparison is across gender. See Table \ref{tab:crosscomparison} for exact numerical values}
\label{fig:crossgendercelebdf}
    \end{minipage}\hfill
    \begin{minipage}{0.45\textwidth}
        \centering
\begin{tikzpicture}
\begin{axis}[
    grid=major, 
    grid style={line width=.2pt, 
    draw=gray!50}, 
    ylabel={Accuracy (\%)},
    xtick=\empty,
    ytick={0,10,20,30,40,50,60,70,80,90,100},
    ymin=10,
    ymax=90,
    bar width=0.2,
    width=7cm,
    height=4cm,
    xtick={0.3,1.5,2.7},
    xticklabels={ {\footnotesize Xception },{\footnotesize Lin et al. },{\footnotesize Ours}},
    tick style={draw=none},
    legend style={
        draw = none,
        at={(0.5,-0.15)},
        anchor=north,
        legend columns=-1,
    },
    enlarge x limits=0.35
]
\addplot[
    style={darkbrown, fill=darkbrown, opacity=1.0},
    ybar=2*\pgflinewidth,
    bar shift=-0.01cm
] 
coordinates {(0,63.99) (1.2,64.10) (2.4,69.61) };

\addplot[
    style={yellow, fill=yellow, opacity=0.6},
    ybar=2*\pgflinewidth,
    bar shift=0.01cm
] 
coordinates {(0.2,81.38)(1.4,80.43) (2.6,81.15) };

\addplot[
    style={purple, fill=purple, opacity=0.9},
    ybar=2*\pgflinewidth,
    bar shift=0.01cm
] 
coordinates {(0.4,17.22)(1.6,32.78) (2.8,70.00) };

\legend{Black, White, Others}

\end{axis}
\end{tikzpicture}
\caption{Cross-dataset evaluation across different methods. Methods are trained on (FF++ \cite{ff++}) and tested on Celeb-DF \cite{celebdf}. Performance comparison is across races. See Table \ref{tab:crosscomparison} for exact numerical values}
\label{fig:crossracecelebdf}
    \end{minipage}
    \label{fig:cross_celebdf}
\end{figure*}

We conducted cross-dataset evaluations to assess our model's performance on out-of-distribution data, evaluating its ability to generalize to unseen datasets. As recorded in Table \ref{tab:crosscomparison}, and visualized in Figures \ref{fig:crossgenderdfdc},\ref{fig:crossracedfdc},\ref{fig:crossgendercelebdf} and \ref{fig:crossracecelebdf}, our approach achieves comparable performance to baseline methods in deepfake detection while significantly outperforming them in fairness generalization across demographic groups.

Figures \ref{fig:crossgenderdfdc} and \ref{fig:crossracedfdc} visualize the results of evaluations from Table \ref{tab:crosscomparison} where the model was trained on FF++ and tested on DFDC, while Figures \ref{fig:crossgendercelebdf} and \ref{fig:crossracecelebdf} visualizes the results from Table \ref{tab:crosscomparison} for training on FF++ and testing on Celeb-DF. These figures demonstrate that our approach exhibits minimal performance disparities across different gender and racial groups compared to the baselines. 

As shown in Table \ref{tab:crosscomparison}, on the DFDC dataset, the Xception baseline has the lowest disparity in accuracy across genders. The disparity in accuracy between males and females is 1.85\% for Xception, and then 4.5\% \cite{fairgen} and 8.03\% (Ours). Across race on the DFDC dataset, our method has the lowest accuracy disparity across the races at 11.15\% (which happens to be between Asian and white) with the baseline recording accuracy disparities of 24.53\% and 14.39\%.

As shown in Table \ref{tab:crosscomparison}, on the Celeb-DF dataset, our method has a minimal disparity in accuracy across both gender and race. Across gender, we got 15.47\% disparity in accuracy, while the baselines recorded wider disparities of 40.66\% and 25.47\% in accuracy. Across race, our method recorded 11.54\% as the maximum disparity in accuracy (which happens to be between white and black), the baselines recorded wider disparities of 26.41\% and 14.39\%.

These results highlight the effectiveness of our approach in achieving fairness generalization in cross-dataset scenarios, demonstrating improved consistency across both racial and gender demographics compared to baseline methods.

\section{Limitations}
\label{sec:limitations}
The limitations of our method are:
\begin{itemize}
\item {Dependence on Demographically Annotated Datasets: } The effectiveness of our method relies on the availability of detailed demographic annotations. Such datasets are often difficult to obtain and may not always be comprehensive enough, potentially impacting the fairness assessment.
\item {Trade-Off Between Fairness and Detection Performance:} Improving fairness across demographic groups can result in a trade-off with overall detection performance. Enhancements aimed at reducing demographic disparities may lead to a decrease in the model's overall accuracy in detecting deepfakes.
\end{itemize}
 
These limitations highlight the need for further research to address the challenges of dataset dependency and the balance between fairness and detection performance.

\section{Conclusion}
\label{sec:conclusion}
We have presented a novel approach to address the critical challenge of fairness generalization in deepfake detection systems. Our method combines three key innovations: a data-centric strategy using synthetic image generation to balance demographic representation, a multi-task learning architecture that simultaneously optimizes detection accuracy and demographic fairness, and sharpness-aware optimization to enhance generalization capabilities.
Our comprehensive evaluations demonstrate the effectiveness of this approach. In intra-dataset testing, our method achieved comparable detection performance to existing approaches while significantly reducing demographic disparities - showing only 0.12\% accuracy difference between gender groups compared to up to 3.87\% in baseline methods, and 0.71\% difference across racial groups compared to up to 5.28\% in baselines. More importantly, in challenging cross-dataset scenarios, our approach demonstrated superior fairness generalization. When tested on the Celeb-DF dataset, our method reduced gender-based accuracy disparities to 15.47\% compared to up to 40.66\% in baselines, while maintaining strong overall detection performance.
These results suggest that our integrated approach of balanced synthetic data, demographic-aware learning, and robust optimization provides a promising direction for developing deepfake detection systems that are both accurate and demographically fair. Future work could explore extending these techniques to other domains where algorithmic fairness and generalization are crucial concerns.

\bibliographystyle{apalike}
{\small
\bibliography{mainbib}}

\end{document}